\documentclass[letterpaper]{article} %
\usepackage{aaai2026}  %
\copyrighttext{%
  \phantom{Copyright \copyright\ 2026, Association for the Advancement of Artificial} \\
  \phantom{Intelligence (www.aaai.org). All rights reserved.}%
}
\usepackage{times}  %
\usepackage{helvet}  %
\usepackage{courier}  %
\usepackage[hyphens]{url}  %
\usepackage{graphicx} %
\usepackage{natbib}  %
\usepackage{caption} %
\usepackage{algorithm}
\usepackage{algorithmic}

\usepackage{newfloat}
\usepackage{listings}
\DeclareCaptionStyle{ruled}{labelfont=normalfont,labelsep=colon,strut=off} %
\floatstyle{ruled}
\newfloat{listing}{tb}{lst}{}
\floatname{listing}{Listing}
\usepackage{xspace}
\usepackage{paralist}

\usepackage{tabularx}
\usepackage{booktabs}

\usepackage[table]{xcolor}  %
\definecolor{indigo}{RGB}{75,0,130}

\newcommand{\martinG}[1]{\textbf{\textcolor{magenta}{Martin G.: #1}}}
\newcommand{\martinA}[1]{\textbf{\textcolor{cyan}{Martin A.: #1}}}
\newcommand{\marc}[1]{\textbf{\textcolor{orange}{Marc: #1}}}
\newcommand{\zongyao}[1]{\textbf{\textcolor{violet}{Zongyao: #1}}}
\newcommand{\marian}[1]{\textbf{\textcolor{teal}{Marian: #1}}}
\newcommand{\oscar}[1]{\textbf{\textcolor{blue}{Oscar: #1}}}
\newcommand{\sebastian}[1]{\textbf{\textcolor{brown}{Sebastian: #1}}}
\newcommand{\johannes}[1]{\textbf{\textcolor{red}{Johannes: #1}}}
\newcommand{\alexs}[1]{\textbf{\textcolor{gray}{Alex S.: #1}}}
\newcommand{\felix}[1]{\textbf{\textcolor{purple}{Felix: #1}}}
\newcommand{\lennart}[1]{\textbf{\textcolor{olive}{Lennart: #1}}}
\newcommand{\joachim}[1]{\textbf{\textcolor{blue}{Joachim: #1}}}
\newcommand{\all}[1]{\textbf{\textcolor{green}{All: #1}}}
\newcommand{\malik}[1]{\textbf{\textcolor{indigo}{Malik: #1}}}

 \renewcommand{\martinG}[1]{}
 \renewcommand{\martinA}[1]{}
 \renewcommand{\marc}[1]{}
 \renewcommand{\zongyao}[1]{}
 \renewcommand{\marian}[1]{}
 \renewcommand{\oscar}[1]{}
 \renewcommand{\sebastian}[1]{}
 \renewcommand{\johannes}[1]{}
 \renewcommand{\alexs}[1]{}
 \renewcommand{\felix}[1]{}
 \renewcommand{\lennart}[1]{}
 \renewcommand{\joachim}[1]{}
 \renewcommand{\all}[1]{}
 \renewcommand{\malik}[1]{}

\newcommand{\router}{\textit{router }}
\newcommand{\chatbot}{\textit{chatbot }}
\newcommand{\pae}{\textit{planner and executor }}
\newcommand{\gcc}{\textit{goal completion critic }}

\title{Agentic AI for Robot Control: Flexible but still Fragile}

\author{
    Oscar Lima\textsuperscript{\rm 1,\rm 2},
    Marc Vinci\textsuperscript{\rm 1,\rm 2},
    Martin Günther\textsuperscript{\rm 1},
    Marian Renz\textsuperscript{\rm 1,\rm 2},
    Alexander Sung\textsuperscript{\rm 1},\\
    Sebastian Stock\textsuperscript{\rm 1},
    Johannes Brust\textsuperscript{\rm 1},
    Lennart Niecksch\textsuperscript{\rm 1,\rm 2},
    Zongyao Yi\textsuperscript{\rm 1,\rm 2},\\
    Felix Igelbrink\textsuperscript{\rm 1,\rm 2},
    Benjamin Kisliuk\textsuperscript{\rm 1,\rm 2},
    Martin Atzmueller\textsuperscript{\rm 2,\rm 1},
    Joachim Hertzberg\textsuperscript{\rm 2,\rm 1}
}
\affiliations{
    \textsuperscript{\rm 1}German Research Center for Artificial Intelligence (DFKI),\\ Cooperative and Autonomous Systems (CAS), Hamburger Straße 24, Osnabrück, Germany\\
    \textsuperscript{\rm 2}Osnabrück University, Institute of Computer Science, Wachsbleiche 27, Osnabrück, Germany\\
    \{oscar.lima, marc.vinci, martin.guenther, marian.renz, alexander.sung, sebastian.stock, johannes.brust, lennart.niecksch, zongyao.yi, felix.igelbrink, benjamin.kisliuk\}@dfki.de\\
    \{martin.atzmueller, joachim.hertzberg\}@uos.de
}

\begin{document}

\maketitle

\begin{abstract}

Recent work leverages the capabilities and commonsense priors of generative models for robot control. In this paper, we present an agentic control system in which a reasoning-capable language model plans and executes tasks by selecting and invoking robot skills within an iterative \pae{} loop. We deploy the system on two physical robot platforms in two settings: (i) tabletop grasping, placement, and box insertion in indoor mobile manipulation (Mobipick) and (ii) autonomous agricultural navigation and sensing (Valdemar). Both settings involve uncertainty, partial observability, sensor noise, and ambiguous natural-language commands. The system exposes structured introspection of its planning and decision process, reacts to exogenous events via explicit event checks, and supports operator interventions that modify or redirect ongoing execution. Across both platforms, our proof-of-concept experiments reveal substantial fragility, including non-deterministic suboptimal behaviour, instruction-following errors, and high sensitivity to prompt specification. %
At the same time, the architecture is flexible: transfer to a different robot and task domain largely required updating the system prompt (domain model, affordances, and action catalogue) and re-binding the same tool interface to the platform-specific skill API.
\end{abstract}

\begin{links}
    \link{Code, prompts and video}{https://dfki-ni.github.io/AGENTS-MAKE-2026}
\end{links}

\section{Introduction}
\label{sec:introduction}

The long-standing AI intuition that a plan is the part of a robot program whose future execution the robot reasons about explicitly~\cite{mcdermott:92} remains a useful organising principle. Yet, bridging abstract plans---readily produced by modern planning systems~\cite{Gha2025}---to real-time physical robot control remains difficult. The core challenge is not nominal execution monitoring, but robustly maintaining and updating the beliefs required for plan-following under partial observability, sensor noise, stale or brittle state estimates, and exogenous dynamics that are not captured by the abstract domain model. Consequently, detecting, interpreting, and responding to unexpected events while remaining plan-consistent and safe is still rather hard in practice.

In this paper, we revisit plan-based robot control through an agentic use of large language models (LLMs), using a single \pae loop that interleaves (i) plan construction in natural language, (ii) tool-grounded execution, and (iii) execution monitoring and recovery. The prompt encodes domain knowledge, axioms, and operational constraints, and the agent queries a sensor-driven semantic state snapshot to ground its decisions under uncertainty. The same loop supports spoken interaction for explanation and for updating task-relevant information during execution. In contrast to end-to-end approaches that map language and raw perception directly to low-level control, we deliberately operate over a higher-level action and perception interface, so that decisions are expressed in terms of interpretable robot skills and verifiable state predicates.

\begin{figure}[t]
    \centering
    \includegraphics[width=0.88\linewidth,angle=-90,origin=c]{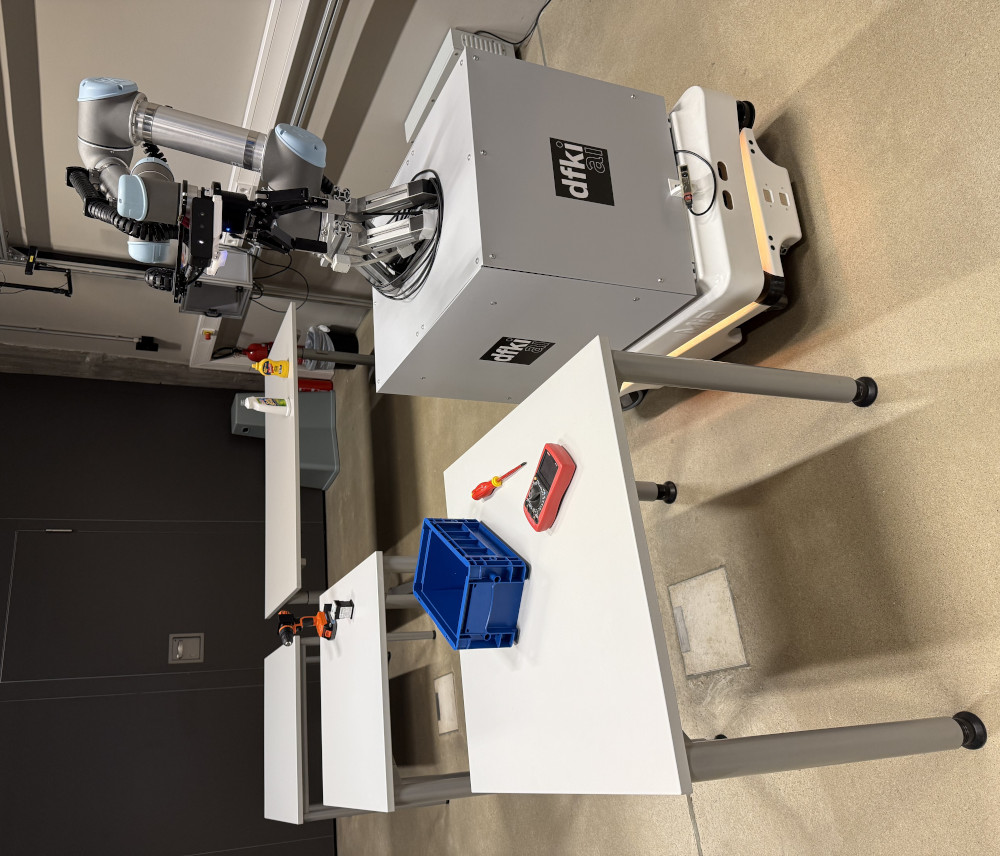}

    \vspace{0.6em}

    \includegraphics[width=0.93\linewidth]{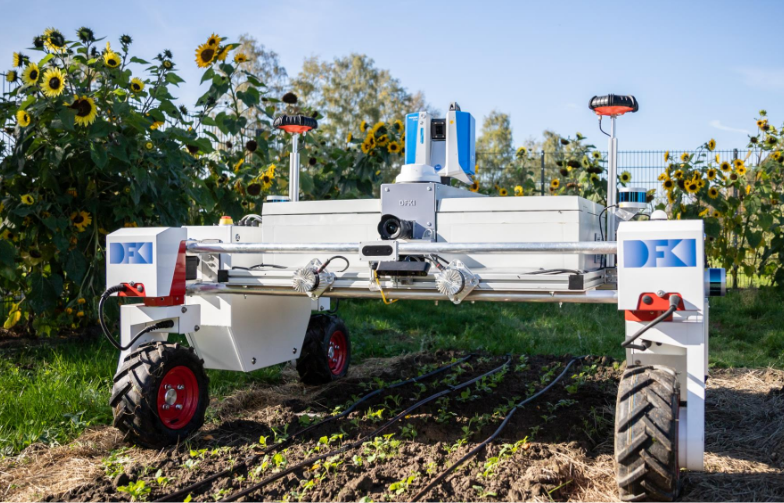}
    \caption{Robotic platforms used in our experiments (see Section~\ref{sec:experimental_evaluation}): the Mobipick mobile manipulator (top), operating in the object collection scenario, and the Valdemar robot (bottom), operating in an agricultural environment.}
    \label{fig:mobipick_and_valdemar}
\end{figure}

We evaluate our proposed approach on two physical robot platforms, Mobipick and Valdemar (Fig.~\ref{fig:mobipick_and_valdemar}). To support partial reproducibility and community access, we also provide a Gazebo-based simulator for the Mobipick scenario, which enables execution of representative tasks despite not reproducing real-world hardware effects; it is available at \url{https://pypi.org/project/mobipick-labs-docker-gui}. Across both platforms, the agent executes tool-grounded plans and closes the loop via verification against the robot's semantic state; when actions fail, it performs consistency checks and replans from structured execution feedback and refreshed state snapshots. We qualitatively validate the workflow on an indoor mobile manipulator in a partially observable tabletop scenario, and report transfer to an outdoor waypoint scanning task in simulation.

\noindent This paper makes the following contributions:
\begin{itemize}
  \item A real-world agentic integration on physical robot platforms, in which a \pae translates natural-language goals into sequences of tool calls over an existing robot stack, with iterative recovery under partial observability and intermittent execution failures.
  \item An architecture-level design that separates geometric and spatial grounding (perception, mapping, localisation, motion and grasp planning) from language-based deliberation: the LLM reasons over structured state snapshots and parametrised atomic skills rather than raw geometry.
  \item A reproducible engineering specification in the form of concrete prompt structure and tool contracts (domain model, affordances, event checking, and recovery patterns), together with a discussion of observed behaviours and fragility, including failure modes arising during long-horizon real-world execution.
\end{itemize}

\subsection{Outline of the Paper}
In Section~\ref{sec:related_work} we summarise relevant prior work, before introducing the proposed agentic control architecture in Section~\ref{sec:agentic_control_architecture}. After that, Section~\ref{sec:modelling} analyses the \pae{} prompt structure, and Section~\ref{sec:agentic_planning_exec} details the corresponding execution pipeline together with the minimal prerequisites for deployment on physical robot platforms. Next, Section~\ref{sec:experimental_evaluation} reports qualitative validation on the physical platforms. Finally, Section~\ref{sec:conclusions} concludes with a summary and discusses interesting directions for future work.

\section{Related Work}
\label{sec:related_work}

Leveraging the common sense and natural language understanding of LLMs for unstructured and open-vocabulary planning problems has been investigated since shortly after the modern transformer-based architectures showed their capabilities. \cite{Jan2020a} uses a fine-tuned GPT-2 model for generating action sequences from a task directive and shows improvements over the popular RNN-based architectures for sequence generation.

\cite{Hua2022a, Wei2022c, singh2022progprompt} improve performance of off-the-shelf LLMs for reasoning in complex tasks without fine-tuning by prompting the LLMs to use a structured format. \cite{Yao2023a} combines the structured reasoning with tool calling (\textit{acting}) and applies this approach also to an embodied setting. A similar approach is proposed by \cite{Ahn2022a}, which uses a separately trained value function for estimating action affordances. These approaches ground actions in information about the environment and, by this, reduce the hallucinations of invalid actions.

The value function in \cite{Ahn2022a} estimates affordances from an image representing the current state of the environment. Other approaches also use images to enrich the context with information about the state of the environment for the LLM \cite{Zen2023d, Dri2023a}. Images and text are encoded into the same token space in \cite{Dri2023a}, which requires fine-tuning for the model to be able to process both modalities simultaneously. On the other hand, \cite{Zen2023d} uses multiple pre-trained models mapping different modalities (text, images, sound) to text for achieving multimodal reasoning without fine-tuning. This approach can also be seen as an early form of the now popular Agentic AI, which orchestrates different models and other tools for accomplishing complex tasks.

To further improve LLM-based reasoning for complex tasks, several works propose to combine planning, acting and feedback on the effects of the action on the environment into a closed loop \cite{huang2023inner, Wan2023d, birr2024autogpt, Ram2024}. Most similar to our proposed approach is the work described in \cite{huang2023inner}, which adds scene descriptions and action success feedback to the context of the LLM. The scene descriptions are either generated from visual models (e.g., object detection or Visual Question Answering) or by requesting input from a person.

\section{Agentic Control Architecture}
\label{sec:agentic_control_architecture}

\begin{figure*}[tb]
    \centering
    \includegraphics[width=\textwidth]{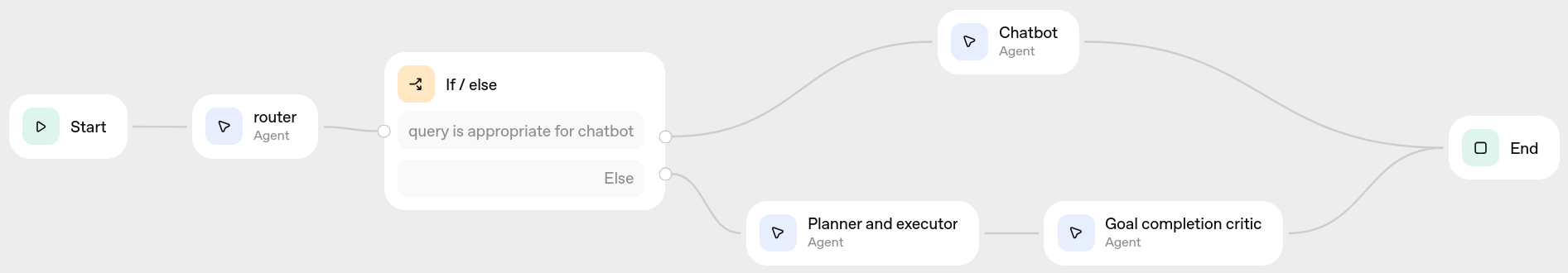}
    \caption{Agent interaction graph (router $\rightarrow$ chatbot or OpenAI o3 based planner/executor $+$ Goal completion critic).}
    \label{fig:multi_agent_topology}
\end{figure*}

Fig. \ref{fig:multi_agent_topology} shows the agent interaction graph used in this paper to perform generative language-based planning and acting for robots. We acknowledge that the present workflow is minimal; a systematic study of more complex designs, including additional agents and alternative communication patterns, is beyond the scope of this paper and left for future work.

The workflow is triggered by a text utterance produced by a speech recognition module. A \router LLM first classifies the utterance as either conversational (question or small talk) or actionable (an instruction requiring execution), and forwards it to the \chatbot or to the \textit{planner and executor}, respectively. The \pae is the primary control agent and uses the OpenAI~o3 reasoning model with a structured prompt that constrains robot behaviour. Using its tools, it can (i) communicate via speech, (ii) query the current sensor driven semantic state, (iii) check for discrete text based events, (iv) invoke environment or state changing actions (navigation, perception, manipulation), and (v) expose introspection via a dedicated function.

Because contemporary tool-using LLMs are trained to repeatedly call external tools until the task is completed, the \pae agent typically continues calling tools until it internally selects a stopping point. In our setup, this termination behaviour is used as is, without additional training or fine tuning. In practice, however, we observed occasional ``verbal only'' failure modes in which the embodied model explains how to solve a task but does not attempt any environment changing action despite feasibility. To mitigate this, we introduce a secondary LLM, the \textit{goal completion critic}, which inspects the execution log (reflections, state, actions, events, and tool results) together with the original user goal. The critic distinguishes between genuine physical attempts (coherent sequences of substantive actions) and superficial conversational output, and can request continuation when the latter occurs.

The \gcc must also handle edge cases in which acting is inappropriate or impossible, for instance when the goal is infeasible, when progress is blocked by repeated physical failures, or when further execution would require explicit user consent. Accordingly, the critic is configured to stop once the system has made a reasonable best effort but becomes blocked, reporting the failure and its cause rather than forcing repeated attempts. In other words, the critic is intended to nudge the \pae away from ``talking through'' feasible tasks, not to drive persistence beyond the point where additional action is unjustified without human intervention.

Table~\ref{tab:agent_configurations} summarises the model assigned to each agent and its configuration, as defined in our agent orchestration setup.\footnote{In our implementation, agent configurations are created with the OpenAI Agent Builder and exported as Python code.} The exported code is then extended with bindings to the robot action APIs (see \cite{lima2023physics} for a full description of the Mobipick robot actions).

\begin{table*}[t]
\centering
\caption{Agent configurations used in the multi agent workflow.}
\label{tab:agent_configurations}
\setlength{\tabcolsep}{5pt}
\renewcommand{\arraystretch}{1.15}
\small
\begin{tabularx}{\textwidth}{@{} l l l l >{\raggedright\arraybackslash}X @{}}
\toprule
\textbf{Agent} & \textbf{Model} & \textbf{Reasoning effort} & \textbf{Output schema} & \textbf{Available tools} \\
\midrule
Router & gpt-5-mini & minimal & bool & None \\
Chatbot & gpt-5-mini & minimal & string &
\texttt{get today date}, \texttt{get available locations}, \texttt{get semantic environment representation snapshot}, \texttt{speak} \\
Planner and executor & o3 & low & string &
\texttt{speak}, \texttt{act}, \texttt{reflect}, \texttt{get semantic environment representation snapshot}, \texttt{check for events} \\
Goal completion critic & o4-mini & low & string & None \\
\bottomrule
\end{tabularx}
\end{table*}

\section{Real-World Robotic Execution: Modelling Task, Environment, and Embodiment}
\label{sec:modelling}

The full prompts for all agents are available on the project website.\footnote{\url{https://dfki-ni.github.io/AGENTS-MAKE-2026/prompts}}
Here, we focus on the overall structure of the \pae agent prompt.

\subsubsection{Planner and Executor Prompt Structure.}

Rather than describing a particular robot stack, the prompt is organised around a set of ingredients we propose as sufficient for real world task execution: an explicit embodiment model, a grounded state acquisition mechanism, a constrained action interface, and norms for safe interaction under partial observability and uncertainty. Concretely, the prompt comprises: (i) a domain model in natural language, (ii) an example semantic environment state, (iii) operational instructions,  (iv) object and action affordances, (v) domain specific heuristics, (vi) the action catalogue, (vii) planning and acting exemplars, and (viii) an optional short memory of prior failure modes to avoid.

\paragraph{Domain Model and State Information.}
The domain model specifies the agent embodiment and its implications for feasibility, sensing, and control. In our tabletop setting, this includes the robot morphology (a mobile base with a single 6 DoF arm and an end effector camera), manipulation constraints (one object at a time; reach limitations), and the sensing semantics (which predicates are observable, and which must be tracked internally). In the agricultural setting, this translated to a robot for agricultural data acquisition, the existence of a shelter with charging station and known locations defined as five points of interest (POI) and the shelter entry position.

The model also encodes that state knowledge is inherently incomplete and time varying: the agent must treat the semantic environment representation as a snapshot that can become stale after motion, human intervention, or perceptual uncertainty. This section therefore defines what must be re-observed (e.g., object pose after navigation) and what cannot be directly sensed (e.g., grasp state), clarifying what information the agent may rely on, guiding subsequent planning. An example illustrating how such a state is provided by the snapshot is also included in the prompt.

\paragraph{Operational Instructions.}
A set of rules is defined that governs execution behavior. The robots require an initial state before acting, periodic state refresh after actions, explicit event handling via a dedicated tool, and a prohibition on inaction when a command remains unfulfilled. We further encode safety-critical constraints that arise from embodiment and operational context, and express them explicitly in the \pae prompt as platform-specific invariants. For the indoor Mobipick demonstrator, this includes retracting the arm before navigation and, when placing, pre-checking available free space on the target surface. For the outdoor Valdemar demonstrator, analogous constraints capture navigation and docking safety, including prohibiting driving while the robot is within the shelter and requiring the robot to be at a designated entry pose before attempting a docking manoeuvre.

\paragraph{Affordances and Heuristics.}
Affordances capture task relevant object properties (for example, heavy tools versus containers) and their interaction consequences (e.g., restrictions on perception poses when carrying certain objects). Heuristics provide bounded, domain grounded preferences that reduce execution risk, such as minimizing the number of actions, delaying placement operations when possible, and preferring information gathering actions when uncertainty is high. Together, these components supply an inductive bias that complements the agent model.

\paragraph{Action Interface and Worked Examples.}
The action catalogue defines the only permitted environment changing operations (e.g., navigation, perception, specific manipulation capabilities, and other platform specific skills). A set of hand crafted planning and acting examples then complement the earlier instructions by steering the \pae towards the intended interaction loop: first produce a brief high level plan in a structured ``reflection'' style (guided by few shot examples but primarily driven by the model's common sense reasoning), then acquire the current state, execute a single grounded action, and check for exogenous events. These working examples also illustrate grounding from language to symbols (e.g., alias resolution to canonical object identifiers) and ambiguity management, either by requesting clarification or by structured exploration when feasible.

\paragraph{Execution Failure Patterns.}
The prompt concludes with a compact, hand curated record of previously observed failure cases, expressed as brief self-criticism that describe what went wrong and what should be done differently next time. Rather than formal prohibitions, these entries function as experience like reminders that bias the \pae away from repeating specific behaviours (e.g., calling the event check repeatedly, or proceeding with placement after recognising that the state is stale). This section targets recurring execution errors and aims to reduce their incidence during long horizon operation.

This prompt structure is intentionally modular: transferring the \pae to a different task domain or robot platform primarily requires updating the domain model, affordances, and action catalogue, while preserving the same interaction loop and safety oriented instructions.

\section{Agentic Planning, Execution, Monitoring, and Recovery from Real-World Failures}
\label{sec:agentic_planning_exec}

\subsection{Planner and Executor Tool Interface}
In this subsection, we describe the tools exposed to the \pae agent exclusively. As summarised in Table~\ref{tab:agent_configurations}, the interface comprises a single environment changing action primitive, plus two state acquisition primitives and one introspection primitive:
\begin{itemize}
  \item \texttt{reflect}: emit structured introspection for debugging and evaluation, without changing the environment.
  \item \texttt{act}: execute exactly one high level action in the environment.
  \item \texttt{get semantic environment representation snapshot}: return the current sensor driven semantic environment representation as a partially observable state snapshot.
  \item \texttt{check\_events}: query for discrete, text based exogenous events since the last check.
\end{itemize}

In our implementation, \texttt{speak} is not a separate tool call but an \texttt{act} action mode. Concretely, \texttt{act} takes an action label and an ordered list of string parameters, constrained by a strict schema (no additional fields). For readability we present it as pseudocode:
\begin{quote}
\textbf{Tool:} \texttt{act(action, params)}\\
\textbf{action} $\in \{\texttt{navigate}, \texttt{move\_arm}, \texttt{perceive}, \texttt{pick},\\ \texttt{place}, \texttt{insert}, \texttt{listen}, \texttt{speak}\}$\\
\textbf{params:} list of strings, interpreted positionally according to \texttt{action}
\end{quote}
This design keeps the tool surface minimal while still supporting dialogue, navigation, perception, and manipulation through a single grounded execution channel.

Execution is delegated to concrete robot skills exposed through a robot specific API, described for the Mobipick platform in \cite{lima2023physics}. Each skill invocation returns a textual status that is interpreted by the \pae (e.g., success, failure, no events, completed speech, or a concise error trace). This explicit feedback channel is critical for faithful failure attribution: without it, the language model may confabulate plausible but incorrect causes. To support correct parameterisation, the \pae has access to skill docstrings that specify expected parameters as well as salient preconditions and effects. During acting, exceptions arising from invalid parameters are propagated to enable immediate correction and retry, while execution failures are extracted from application logs and returned as succinct diagnostics (e.g., infeasible grasp or unreachable pose). Together with \texttt{reflect} outputs and \texttt{semantic environment representation snapshot} queries, these outcomes provide the evidence the \pae uses to revise its plan, when new information becomes available, even though replanning is not hard coded in the bindings but emerges from the interaction loop. Empirically, when evidence invalidates earlier assumptions, the agent typically adapts without being destabilised by stale intermediate history, including over long interaction traces as shown in our proof of concept demonstrations.

\subsection{Reflection as ``Planning''}
As noted in Section~\ref{sec:related_work}, the use of an explicit self-authored reasoning trace is not new. Huang et~al. introduce a closely related mechanism under the name \emph{Inner Monologue} \cite{huang2023inner}, which serves the same functional role: externalising intermediate reasoning to support multi-step decision making. Our contribution is to revisit this pattern in the context of contemporary agentic tool use, where recent reasoning oriented language models are more capable of maintaining goal directed plans while interleaving deliberation with action execution. Despite still exhibiting limitations in systematic planning performance and scaling to larger problem instances~\cite{valmeekam2025systematic}, these models can be sufficient for small, structured robotics domains. Crucially, they are trained to operate via tool invocation, making the reflection channel a practical planning interface for an agent that must iteratively sense, decide, and act in the physical world.

\subsection{Real-World Execution Prerequisites}
Our approach assumes access to a robot action API that exposes a set of executable skills (for example, navigation, observation, grasping, placement, and insertion). In the indoor implementation, these skills are backed by a complete robot software stack built on ROS~1 and described in detail in \cite{lima2023physics}. Developed separately but following the same concepts, the agricultural robot also provides a Python API that encapsulates the underlying ROS~1 calls.

Concretely, in the indoor mobile manipulation demonstrator (Mobipick), execution requires autonomous navigation, object detection with classification and six degree of freedom pose estimation, grasp synthesis, motion planning via MoveIt, inverse kinematics, and trajectory execution with monitoring for manipulation actions. Placement further requires sampling and feasibility checking of candidate place poses, again as detailed in \cite{lima2023physics}. For other platforms and task domains (for example, outdoor navigation and sensing on the agricultural robot), the specific low level components differ, but the architectural requirement remains an action API exposing executable skills
and returning structured success or failure signals. While our experiments rely on classical robot control components, the same interface can wrap higher level robot foundation model capabilities as individual skills or skill bundles (e.g., loading a dishwasher or folding laundry), provided they are callable through an action API and return structured success or failure signals.

To support planning under partial observability, the \pae consumes a semantic environment representation as a compact state snapshot. In the indoor system, this representation is generated by a lightweight symbolic fact generation pipeline\footnote{\url{https://github.com/DFKI-NI/symbolic_fact_generation}}. This pipeline is described in detailed in \cite{lima2023physics}. It is a collection of Python utilities that transform buffered perception outputs into discrete predicates, such as \texttt{at(robot, location)}, \texttt{on(object, table\_1)}, \texttt{in(object, box)}, and \texttt{arm\_posture(robot, observe)}. A key mechanism is an observation buffer populated by the perception module whenever the end effector camera is directed at a workspace via a dedicated arm motion to a recorded observation posture (implemented as a fixed joint configuration).
In the agricultural robot system, an analogous abstraction is constructed from navigation and platform telemetry. Concretely, the robot's current geoposition is mapped to the nearest known location in the environment description, or to \textit{unknown} when it is outside the described region. Likewise, the battery state of charge is discretised using preset thresholds into \textit{okay}, \textit{low}, or \textit{critical}. These discrete, natural language like assertions are then returned to the agent through the same channel, enabling uniform downstream reasoning despite differing sensors and operating conditions.
The resulting state abstraction is intentionally simple in both settings and can be replaced by richer semantic mapping, for example using approaches in the spirit of \cite{renz2026expris}, without changing the agent interaction loop.

\section{Qualitative Validation on a Physical Robot with Simulated Transfer}
\label{sec:experimental_evaluation}

This section presents a set of experiments designed to validate the generality and adaptability of the agentic system across heterogeneous robotic platforms and task domains. In particular, we demonstrate that the same underlying architecture can be deployed with only minimal, task-specific adaptations on both indoor and outdoor mobile robotic systems. The experiments were intentionally designed as proof-of-concept demonstrations rather than exhaustive quantitative evaluations. The primary objective was to evaluate whether the proposed system functions as intended and can be transferred across platforms and domains with only minimal, local modifications.

The system was first validated on the indoor mobile manipulator in both simulation and the real-world setup. Subsequently, within approximately one day of development effort, the same system was adapted to the outdoor agricultural robot's API and tested in simulation.

\subsection{Qualitative Validation Setup}
Two robotic demonstrators were used in the experiments (see Fig.~\ref{fig:mobipick_and_valdemar}). The indoor mobile manipulator Mobipick combines a mobile base with a robotic arm and gripper. The outdoor agricultural mobile robot called Valdemar is intended for garden and crop monitoring.

\subsubsection{Indoor Scenario Description}

The indoor experiments were conducted in a partially observable environment consisting of three tables, as shown in Fig.~\ref{fig:mobipick_and_valdemar}. A set of objects is distributed across the tables. The objects used in the experiments include a screwdriver, a relay, a multimeter, a power drill, and a blue box serving as a container for transporting the other smaller objects. A bleach bottle is also present and can be interacted with, but was not explicitly named in the domain prompt beforehand.

The robot is required to perform a range of manipulation actions, including picking up objects, placing them on tables, and inserting them into containers. Due to partial observability, the robot does not have complete knowledge of the initial object locations and must actively perceive and search for objects before manipulating them. This setting emphasizes the need for tight integration between perception, decision-making, and action execution.

\subsubsection{Outdoor Scenario Description}
The outdoor experiments focus on navigation and sensing tasks in a garden-like environment under the influence of external dynamics, exemplified by the battery management. The agricultural robot is tasked with autonomously navigating to a sequence of predefined waypoints and performing plant scanning operations at each location. To avoid the battery being depleted, the robot must continuously monitor its battery state. Thus, it can ensure sufficient energy or interrupt operation to recharge the batteries.

This scenario introduces different operational requirements compared to the indoor setup, notably long-range navigation, environmental variability, and resource-aware decision-making, while still relying on the same underlying system.

\subsection{Experiments}

\subsubsection{Experiment 1 (Mobipick): Nominal Task Execution}
In the first experiment, the robot was instructed to \emph{``put the multimeter inside the box and bring it to table~2''}. The task required reasoning about object locations, container use, and a target placement location under partial observability.

Across most runs, the system produced a consistent plan structure: it searched for the multimeter, picked it up after detection, searched for the blue box, inserted the multimeter into the box, and transported the box to the specified table. This demonstrated basic integration of perception, symbolic reasoning, and manipulation.
The runs also exposed limitations of the current implementation. In particular, placement was occasionally executed without verifying free space on the target table, resulting in ``blind'' placement actions. Some plans were also inefficient; for example, the robot sometimes searched for both the multimeter and the box before returning to pick up the multimeter, introducing avoidable navigation and perception steps. These issues did not typically prevent task completion, but they indicated scope for more informed planning and execution constraints.

\subsubsection{Experiment 2 (Mobipick): Ambiguous Command and Event Handling}
The second experiment examined ambiguity resolution, human--robot interaction, and robustness to execution interruptions. The robot received the ambiguous instruction \emph{``pick any tool I can use to screw with''}. In the environment, both a screwdriver and a power drill satisfied the description, requiring an explicit selection. After perceiving both tools on the table, the system selected the screwdriver based on a commonsense preference for a lighter tool, despite the prompt encoding that the screwdriver was harder to grasp due to its small size. This mismatch between prompt-specified affordances and model reasoning led to a planned picking motion that would have collided with the table, and the user activated the emergency stop.
The triggered emergency stop was provided to the \pae as an unexpected event. After checking for events, the system detected that the arm could no longer be actuated, inferred that execution could not continue, and entered a waiting state without issuing further manipulation actions until the user released the emergency stop. The robot then attempted to resume by picking the screwdriver, but the user intervened and expressed a preference for the power drill. The system incorporated this update via a further event check, revised its plan, abandoned the screwdriver, and successfully picked the power drill.

Overall, the episode showed that the system handled underspecified instructions, responded appropriately to safety interventions, and adapted its behaviour online in response to user feedback.

\subsubsection{Experiment 3 (Mobipick): Constraint-Aware Placement under Partial Perception}
This experiment evaluated the system's ability to reason about domain constraints arising from partial perception and to incorporate these constraints proactively during planning. In particular, the robot was constrained to never place objects on table surfaces that had not been perceived beforehand, in order to avoid collisions with potentially present but unknown objects.
The initial environment setup consisted of a box on table~3 containing a multimeter. The task instruction given to the system was: \emph{``go to table~3 and take the box from there to table~2''}. Due to limitations of the perception action, the robot could not perceive the table surface while holding the box if the box contained an object. Attempting to perceive under this condition required reorienting the arm, which would invert the box and cause its contents to fall out. This constraint was represented in the system as a safety restriction.
Given these constraints, a plan that first picked up the box and subsequently attempted to perceive table~2 would have been unsafe. Correct behaviour thus required anticipatory planning: the robot had to first navigate to table~2 and perceive its surface, and only then proceed to table~3 to pick up the box and transport it directly to the already perceived target surface.

In this experiment, the robot successfully planned and executed the task as required. No collisions occurred, and the multimeter remained inside the box throughout execution, consistent with the specified constraints.

\subsubsection{Experiment 4 (Mobipick): Long-Horizon Stress Test}
This experiment served as a stress test for the system under a long planning horizon with a large number of sequential and interdependent actions. The objective was to probe robustness under extended task execution, including opportunistic tool use and the maintenance of an up to date world model in an increasingly cluttered workspace.

The robot was instructed to \emph{``transport all objects to table~2''}. To improve efficiency, the robot could use a box to transport multiple objects simultaneously. As objects accumulated on table~2, the available placement area decreased, requiring the robot to maintain sufficiently recent perceptual information about object poses to execute safe placements.

In this experiment, the robot collected all available objects from multiple tables and consolidated them on table~2. It used the box as a container to transport multiple items efficiently, reducing the number of transport cycles. When placing the box, the robot did not refresh its perception of table~2 immediately prior to placement, resulting in a comparatively risky placement action; this placement nevertheless succeeded because table~2 still had ample free space at that stage. The bleach bottle, which was part of the environment but not explicitly named in the prompt, was also collected. While grasping the bleach bottle, excessive tension in an arm mounted cable triggered the emergency stop. The system paused, accepted user intervention, and subsequently resumed the task. Finally, the robot navigated back to table~1 to verify that no objects remained. Although this was not strictly necessary for completing the instruction, it reflected a conservative strategy: because a non-negligible amount of time had elapsed, the system treated the table state as potentially outdated and chose to re-verify it.

The experiment demonstrated that the system can handle long sequences of actions, exploit containers to reduce transport overhead, and recover from an execution-level interruption. At the same time, it exposed a weakness in automatic world-state maintenance: object poses on the target table were not proactively refreshed before every placement, and one placement relied on stale information. This observation suggests that additional heuristics or constraints, for example mandatory re-perception under high clutter or after prolonged execution, could improve robustness in long-horizon tasks.

\subsubsection{Experiment 5 (Valdemar): Nominal Task Execution Outdoors}
To demonstrate nominal functioning of the agentic control system, the robot was tasked to \textit{``scan poi\_1, poi\_2, poi\_3, poi\_4, poi\_5''}, matching the exact identifiers of predefined, known locations. This required the robot to determine whether it was docked in the shelter (necessitating undocking before departure) and to verify that the battery status was \texttt{okay} before proceeding. In typical runs, the system queried the current environment state, correctly identified that the robot was docked and that sufficient energy was available, then undocked and navigated to each point of interest, performing a scan at every waypoint. Finally, it returned to the shelter and docked to the charging station, as instructed by the prompts. During our tests of this scenario, no issues were observed.

\subsubsection{Experiment 6 (Valdemar): Low Battery Monitoring}
In this experiment, the robot was instructed to scan three points of interest, inducing substantial travel time. We initialised the simulated battery at a low state of charge and increased the consumption rate parameter such that, during execution, the battery state transitioned to \emph{``low''}. In repeated trials, the \pae interrupted the mission, navigated back to the shelter, docked, and invoked a charging action that restored the battery state to \textit{``okay''}. It then resumed from the previously interrupted point of the route, completed the remaining scans, and finally returned to the shelter.

A further observation is that, in the absence of an explicit recharge target, the \pae selected different charge goals across runs (typically 80\%, 90\%, or 100\%), indicating nondeterminism in the chosen stopping criterion under underspecified objectives.

While the overall behaviour is functional, a limitation is exposed by the event handling mechanism: the system must actively poll for events, implying that execution is only interrupted between discrete actions. Consequently, long running actions are not preempted. For instance, during a navigation action the robot may continue moving away from the shelter even if the battery threshold is crossed shortly after the action begins, postponing corrective behaviour until the action completes.

\subsubsection{Experiment 7 (Valdemar): Invalid Command Refutal}
In this experiment, the robot was given commands which sounded like planning and execution requests, but were impossible by nature, such as \textit{``Fly to Paris''}. The agentic system delegated this request to the \textit{planner and executor}, which refuted the request by reasoning about its static domain knowledge -- \textit{``I can't fly''}, \textit{``Paris it far outside my operation area''} -- and explaining this to the user, using the speak function.

In other cases, where the robot was asked to play a song or execute an impossible movement, the agentic system delegated to the chatbot or refused the command similar to the description above. This raises an interesting point of the API as an abstraction layer between the agent and the low-level control: With model alignment and security against prompt injection being out of scope of this paper, the limitation to valid API calls ensured safe behavior of the robot.

During these experiments, the agentic system showed hallucinative behavior: While it correctly refused to \textit{``do a somersault''}, it offered alternatives like driving patterns which were likewise impossible to achieve, seemingly to please the user.

\subsection{Qualitative Observations}
In practice, enforcing consistent event checking proved difficult. Although the prompt specifies when to invoke the event checking tool and provides illustrative examples, the \pae may call it irregularly, including repeated invocations in rapid succession. This behaviour persisted despite multiple prompt level interventions, highlighting limited controllability of tool use policies under long horizon execution.

On the other hand, the system showed some unexpected beneficial behaviors: During testing, a non-expert user asked the agricultural system to \textit{``demonstrate a real scan''} at which point the system executed a simple charge-undock-navigate-scan-return routine and used the speak action to explain its process, demonstrating the potential for much improved transparency for operators.

\section{Conclusions}
\label{sec:conclusions}

In contrast to classical planning systems that depend on expert crafted planning domains, our approach encodes task and action knowledge in natural language, enabling rapid adaptation to related robotic platforms with modest integration effort.
Beyond describing feasible behaviour, the system can interpret and verbalise its ongoing reasoning and task execution, which can increase transparency of internal state and decision making for non expert end users and operators.

Our architecture assumes an operational robot software stack that exposes a closed set of preprogrammed, parametrised atomic skills through an action API. Under this abstraction, the language model is restricted to high level decision making and tool invocation, while perception and low-level control remain implemented in the underlying stack. In contrast to end to end robot foundation models that jointly learn perception, control, and policy, the present system does not learn new low-level capabilities from data and is therefore applicable when the requisite skills are already available on the platform.

Recent LLMs provide substantially larger context windows than earlier generations, which makes it practical to maintain and condition on rich task context, action specifications, and accumulated execution traces within a single \pae loop. This increase in usable context is a key enabler of the kind of long horizon, language grounded robotic task execution demonstrated in this paper; comparable integrations were markedly less feasible when models supported only short context lengths.

We inherit from the LLMs several practical capabilities that are directly useful for deployed human robot interaction. In our experiments, a single \pae prompt structure supported open vocabulary references and natural language paraphrases of objects, locations, and actions, including multilingual utterances, without requiring changes to the underlying robot skill API. Moreover, encoding object and action affordances at the prompt level provides a lightweight mechanism to inject task relevant priors (e.g., what can be grasped, carried, perceived, or safely placed) into decision making, enabling the \pae to combine these explicit constraints with its broader linguistic and common sense priors when grounding instructions into executable tool calls.

A further set of limitations concerns operational cost, latency, and deployment assumptions. Because the \pae relies on remote model inference in the present implementation, each tool mediated decision step incurs a monetary cost per query and introduces non-negligible end-to-end latency; we do not report measured timings here, and therefore only claim qualitatively that the iterative loop adds delay relative to purely local control.
This deployment choice also implies dependence on a reliable internet connection, which is only indirectly addressed above; if stated explicitly, it should be read together with the future-work mitigation on local, on-device deployment. Finally, our experience indicates that robust domain modelling remains labour intensive in practice: to compensate for prompt sensitivity and limited controllability, safety critical constraints and invariants often need to be stated redundantly across multiple prompt components (domain model, operational rules, exemplars, and failure-pattern reminders), yet occasional constraint violations still occur under long horizon execution.

A further limitation concerns asynchrony and event handling. In our implementation, tool use is synchronous within an iterative loop and exogenous changes are detected via explicit event checks, i.e., polling. Reacting to events ``at any time'' requires (i) concurrent event detection during ongoing work and (ii) an explicit cancellation or preemption interface in the executing operation. Most contemporary LLM agent frameworks do not provide hard preemption of arbitrary tool calls nor a uniform interruption primitive. Therefore, robust interruption of physical execution under asynchronous events generally requires exposing robot skills as cancellable, preemptible tasks and running event monitoring concurrently, or decomposing long actions into checkpointed, interruptible segments, at the cost of additional coordination complexity.

\subsection{Future Work}
We identify several possible future research directions and implementation optimisations. These include exploring more strongly agentic decompositions via additional specialised agents (e.g., separate planner, executor, verification and critique, and history compaction); evaluating alternative inter agent communication patterns beyond the current linear pipeline, such as manager style hierarchical delegation or all to all interaction; assessing local deployment with on device language models to reduce reliance on network connectivity; proposing a simulated physics-based robot execution benchmark to enable systematic comparisons across models and workflow variants; and studying whether reinforcement-based fine tuning (e.g., agent RFT) can improve robustness and latency, potentially reducing dependence on external runtime critics.

\section*{Acknowledgments}
This work is supported by the ExPrIS and LIEREx projects through grants from the German Federal Ministry of Research, Technology and Space (BMFTR) with grant numbers 16IW23001 and 16IW24004.
The DFKI Niedersachsen (DFKI NI) is sponsored by the Ministry of Science and Culture of Lower Saxony and the VolkswagenStiftung.

We thank Malik Ghallab (Directeur de Recherche Emeritus at CNRS and the University of Toulouse) for constructive feedback and for sharing key insights that shaped this work.

\bibliography{References}

@InProceedings{Ahn2022a,
  author    = {Ahn, Michael and Brohan, Anthony and Brown, Noah and Chebotar, Yevgen and Cortes, Omar and David, Byron and Finn, Chelsea and Gopalakrishnan, K. and Hausman, Karol and Herzog, Alexander and Ho, Daniel and Hsu, Jasmine and Ibarz, Julian and Ichter, Brian and Irpan, A. and Jang, Eric and Ruano, Rosario Jauregui and Jeffrey, Kyle and Jesmonth, Sally and Joshi, N. and Julian, Ryan C. and Kalashnikov, Dmitry and Kuang, Yuheng and Lee, Kuang-Huei and Levine, Sergey and Lu, Yao and Luu, Linda and Parada, Carolina and Pastor, Peter and Quiambao, Jornell and Rao, Kanishka and Rettinghouse, Jarek and Reyes, D. and Sermanet, P. and Sievers, Nicolas and Tan, Clayton and Toshev, Alexander and Vanhoucke, Vincent and Xia, F. and Xiao, Ted and Xu, Peng and Xu, Sichun and Yan, Mengyuan},
  title     = {Do As I Can, Not As I Say: Grounding Language in Robotic Affordances},
  booktitle = {Conference on Robot Learning (CoRL)},
  year      = {2022},
  month     = apr,
}

@InProceedings{birr2024autogpt,
  author    = {Timo Birr and Christoph Pohl and Abdelrahman Younes and Tamim Asfour},
  title     = {{AutoGPT+P}: Affordance-based Task Planning using Large Language Models},
  booktitle = {Robotics: Science and Systems (RSS)},
  year      = {2024},
  doi       = {10.15607/RSS.2024.XX.112},
  url       = {https://doi.org/10.15607/RSS.2024.XX.112},
}

@InProceedings{Dri2023a,
  author    = {Driess, Danny and Xia, Fei and Sajjadi, Mehdi S. M. and Lynch, Corey and Chowdhery, Aakanksha and Ichter, Brian and Wahid, Ayzaan and Tompson, Jonathan and Vuong, Quan and Yu, Tianhe and Huang, Wenlong and Chebotar, Yevgen and Sermanet, Pierre and Duckworth, Daniel and Levine, Sergey and Vanhoucke, Vincent and Hausman, Karol and Toussaint, Marc and Greff, Klaus and Zeng, Andy and Mordatch, Igor and Florence, Pete},
  title     = {{PaLM-E}: An Embodied Multimodal Language Model},
  booktitle = {Proceedings of the 40\textsuperscript{th} International Conference on Machine Learning (ICML)},
  year      = {2023},
  volume    = {202},
  pages     = {8469--8488},
  month     = jul,
  publisher = {PMLR},
}

@Book{Gha2025,
  title     = {Acting, Planning, and Learning},
  publisher = {Cambridge University Press},
  year      = {2025},
  author    = {Ghallab, Malik and Nau, Dana and Traverso, Paolo},
}

@InProceedings{Hua2022a,
  author    = {Huang, Wenlong and Abbeel, Pieter and Pathak, Deepak and Mordatch, Igor},
  title     = {Language Models as Zero-Shot Planners: Extracting Actionable Knowledge for Embodied Agents},
  booktitle = {Proceedings of the 39\textsuperscript{th} International Conference on Machine Learning (ICML)},
  year      = {2022},
  pages     = {9118--9147},
  month     = jun,
  publisher = {PMLR},
}

@InProceedings{huang2023inner,
  author    = {Wenlong Huang and Fei Xia and Ted Xiao and Harris Chan and Jacky Liang and Pete Florence and Andy Zeng and Jonathan Tompson and Igor Mordatch and Yevgen Chebotar and Pierre Sermanet and Tomas Jackson and Noah Brown and Linda Luu and Sergey Levine and Karol Hausman and Brian Ichter},
  title     = {Inner Monologue: Embodied Reasoning through Planning with Language Models},
  booktitle = {Proceedings of the 6\textsuperscript{th} Conference on Robot Learning (CoRL)},
  year      = {2023},
  volume    = {205},
  series    = {Proceedings of Machine Learning Research},
  pages     = {1769--1782},
  publisher = {PMLR},
}

@InProceedings{Jan2020a,
  author    = {Jansen, Peter},
  title     = {Visually-Grounded Planning without Vision: Language Models Infer Detailed Plans from High-level Instructions},
  booktitle = {Findings of the Association for Computational Linguistics: EMNLP 2020},
  year      = {2020},
  pages     = {4412--4417},
  month     = nov,
  publisher = {Association for Computational Linguistics},
  doi       = {10.18653/v1/2020.findings-emnlp.395},
}

@InProceedings{lima2023physics,
  author    = {Lima, Oscar and G{\"{u}}nther, Martin and Sung, Alexander and Stock, Sebastian and Vinci, Marc and Smith, Amos and Krause, Jan Christoph and Hertzberg, Joachim},
  title     = {A Physics-Based Simulated Robotics Testbed for Planning and Acting Research},
  booktitle = {ICAPS Workshop on Planning and Robotics (PlanRob 2023)},
  year      = {2023},
  address   = {Prague, Czech Republic},
  month     = jul,
}

@Article{mcdermott:92,
  author  = {D. McDermott},
  title   = {Robot Planning},
  journal = {AI Magazine},
  year    = {1992},
  volume  = {13},
  number  = {2},
  pages   = {55--79},
}

@InProceedings{Ram2024,
  author    = {Raman, Shreyas Sundara and Cohen, Vanya and Idrees, Ifrah and Rosen, Eric and Mooney, Raymond and Tellex, Stefanie and Paulius, David},
  title     = {{CAPE}: Corrective Actions from Precondition Errors Using Large Language Models},
  booktitle = {IEEE International Conference on Robotics and Automation (ICRA)},
  year      = {2024},
  pages     = {14070--14077},
  month     = may,
  doi       = {10.1109/ICRA57147.2024.10611376},
}

@Article{renz2026expris,
  author    = {Renz, Marian and G{\"u}nther, Martin and Igelbrink, Felix and Lima, Oscar and Atzmueller, Martin},
  title     = {{ExPrIS}: Knowledge-Level Expectations as Priors for Object Interpretation from Sensor Data},
  journal   = {KI-K{\"u}nstliche Intelligenz},
  year      = {2026},
  pages     = {1--6},
  publisher = {Springer},
}

@InProceedings{singh2022progprompt,
  author    = {Singh, Ishika and Blukis, Valts and Mousavian, Arsalan and Goyal, Ankit and Xu, Danfei and Tremblay, Jonathan and Fox, Dieter and Thomason, Jesse and Garg, Animesh},
  title     = {{ProgPrompt}: Generating Situated Robot Task Plans Using Large Language Models},
  booktitle = {IEEE International Conference on Robotics and Automation (ICRA)},
  year      = {2023},
  pages     = {11523--11530},
  doi       = {10.1109/ICRA48891.2023.10161317},
}

@Article{valmeekam2025systematic,
  author  = {Valmeekam, Karthik and Stechly, Kaya and Gundawar, Atharva and Kambhampati, Subbarao},
  title   = {A Systematic Evaluation of the Planning and Scheduling Abilities of the Reasoning Model o1},
  journal = {Transactions on Machine Learning Research},
  year    = {2025},
}

@InProceedings{Wan2023d,
  author    = {Wang, Zihao and Cai, Shaofei and Chen, Guanzhou and Liu, Anji and Ma, Xiaojian and Liang, Yitao},
  title     = {Describe, Explain, Plan and Select: Interactive Planning with {LLMs} Enables Open-World Multi-Task Agents},
  booktitle = {Advances in Neural Information Processing Systems (NeurIPS)},
  year      = {2023},
  volume    = {36},
  pages     = {34153--34189},
  publisher = {Curran Associates, Inc.},
}

@InProceedings{Wei2022c,
  author    = {Wei, Jason and Wang, Xuezhi and Schuurmans, Dale and Bosma, Maarten and Ichter, Brian and Xia, Fei and Chi, Ed H. and Le, Quoc V. and Zhou, Denny},
  title     = {Chain-of-Thought Prompting Elicits Reasoning in Large Language Models},
  booktitle = {Advances in Neural Information Processing Systems (NeurIPS)},
  year      = {2022},
  publisher = {Curran Associates Inc.},
}

@InProceedings{Yao2023a,
  author    = {Yao, Shunyu and Zhao, Jeffrey and Yu, Dian and Du, Nan and Shafran, Izhak and Narasimhan, Karthik R and Cao, Yuan},
  title     = {{ReAct}: Synergizing Reasoning and Acting in Language Models},
  booktitle = {International Conference on Learning Representations (ICLR)},
  year      = {2023},
}

@InProceedings{Zen2023d,
  author    = {Zeng, Andy and Attarian, Maria and Ichter, Brian and Choromanski, Krzysztof Marcin and Wong, Adrian and Welker, Stefan and Tombari, Federico and Purohit, Aveek and Ryoo, Michael S and Sindhwani, Vikas and Lee, Johnny and Vanhoucke, Vincent and Florence, Pete},
  title     = {Socratic Models: Composing Zero-Shot Multimodal Reasoning with Language},
  booktitle = {International Conference on Learning Representations (ICLR)},
  year      = {2023},
}

\end{document}